\pdfoutput=1

\documentclass[11pt]{article}

\usepackage{acl} 

\usepackage{times}
\usepackage{latexsym}
\usepackage{color,colortbl}
\usepackage{graphicx,subcaption}
\usepackage[normalem]{ulem}
\usepackage{multirow}
\usepackage{xspace,mfirstuc,tabulary}

\usepackage{amsmath} 
\useunder{\uline}{\ul}{}
\usepackage[T1]{fontenc}
\usepackage{microtype}
\usepackage[utf8]{inputenc}

\usepackage{microtype}

\usepackage{csquotes}

\title{Through the Lens of Split Vote: Exploring Disagreement, Difficulty and Calibration in Legal Case Outcome Classification}

\author{Shanshan Xu$^{1,2*}$, Santosh T.Y.S.S$^{1}$,Oana Ichim$^{3}$ \\
\textbf{Barbara Plank$^{4,5}$, Matthias Grabmair$^1$}\\
$^{1}$Technical University of Munich, Germany  $^{2}$ELTEMATE\\  
$^{3}$Graduate Institute of International and Development Studies, Switzerland\\ 
$^{4}$IT University of Copenhagen, Denmark\\
$^{5}$LMU Munich \& Munich Center for Machine Learning (MCML), Germany\\
}

\begin{document}
\maketitle

\begin{abstract}

In legal decisions, split votes (SV) occur when judges cannot reach a unanimous decision, posing a difficulty for lawyers who must navigate diverse legal arguments and opinions. 
In high-stakes domains, 
understanding the alignment of perceived difficulty between humans and AI systems is crucial to build trust. However, existing NLP calibration methods focus on a classifier's awareness of predictive performance, measured against the human majority class, overlooking inherent human label variation (HLV). This paper explores split votes as naturally observable 
human disagreement and value pluralism. We collect judges’ vote distributions from the European Court of Human Rights (ECHR), and present \textbf{SV-ECHR}\footnote{Our dataset and code is available at \url{https://github.com/TUMLegalTech/SplitVote\_ECHR}} a case outcome classification (COC) dataset with SV information. We build a taxonomy of disagreement with SV-specific subcategories. We further assess the alignment of perceived difficulty between models and humans, as well as confidence- and human-calibration of COC models. We observe limited alignment with the judge vote distribution. To our knowledge, this is the first systematic exploration of calibration to human judgements in legal NLP. Our study underscores the necessity for further research on measuring and enhancing model calibration considering HLV in legal decision tasks.

\end{abstract}

\def\thefootnote{*}\footnotetext{Work done during internship at ELTEMATE}

\section{Introduction}  

\begin{figure}[t]
    \centering
    \includegraphics[width = 0.49\textwidth]
    {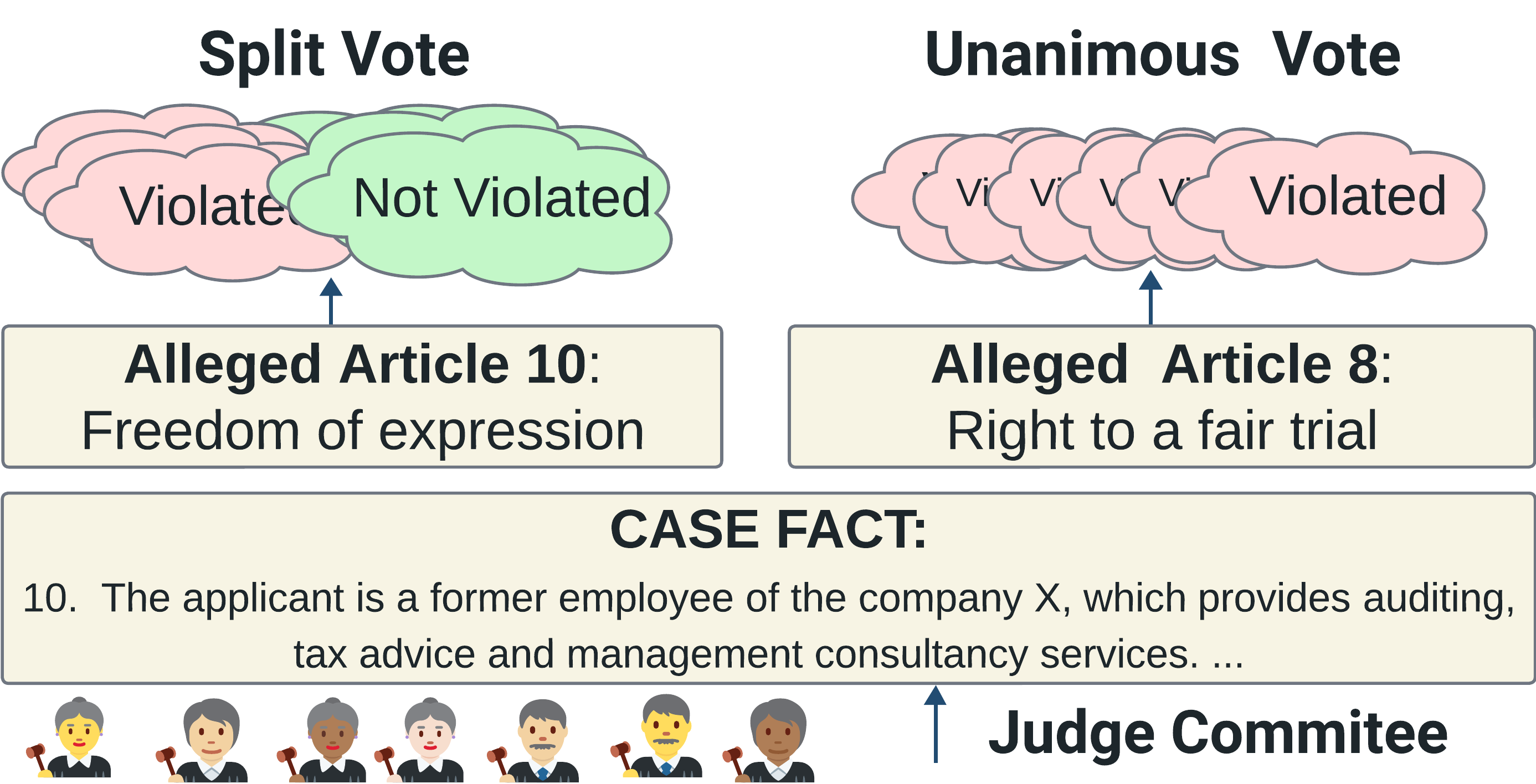}
    \vspace{-0.1cm}
    \caption{Split Votes in ECtHR Decisions}
    \label{fig:fig1}
\end{figure}



The task of Case Outcome Classification (COC) involves the classification of legal case outcomes based on textual descriptions of their facts. While achieving high performance is desirable, in high-stakes domains such as legal and medical decisions, the quantification of a model's predictive confidence, or conversely, its uncertainty, is particularly valuable. It allows experts to make more informed decisions, especially when the model may be uncertain or where the consequences of a misdiagnosis are significant. Evaluating whether models are aware of their limitations is known as assessing their uncertainty, with popular methods such as \textit{calibration} \cite{guo2017calibration,desai-durrett-2020-calibration}. Calibration assesses the extent to which predictive probabilities accurately reflect the likelihood of a prediction being correct. Models can opt to abstain when the uncertainty exceeds a predefined threshold — a method commonly referred to as \textit{selective classification}         \cite{el2010foundations,geifman2017selective}.

Current NLP research focuses on prediction confidence and calibration to assess a classifier's awareness of its predictive performance only. This evaluation is commonly conducted against the human majority class.
However, recent developments in NLP research have shed light on the prevalence of inherent human label variation (HLV)~\cite{plank-2022-problem}, observing disagreement\footnote{The term HLV embraces disagreement and plausible variation. In this paper, we  use these terms interchangeably.} across various tasks \cite{uma2021learning}. Scholars in the field argue for the acknowledgment and acceptance of HLV, as it mirrors the diverse and pluralistic nature of human values \cite{sorensen2024roadmap}. Notably, \citealt{baan-etal-2022-stop} has demonstrated that widely-used calibration metrics may not be suitable for datasets characterized by inherent human disagreement.

In light of these findings, we posit that models should not only be calibrated to recognize their own uncertainty (performance calibration) but also be equipped to discern instances where humans exhibit uncertainty (human calibration) \cite{baan-etal-2022-stop}. This dual focus aims to foster trust among end-users and mitigates potential harm caused by models. Consequently, a critical aspect of this trust involves ensuring the alignment of perceived difficulty between human and models.

This motivates us to study split votes (SV) in court decisions (\autoref{fig:fig1}). 
The judge vote ratio is a \textit{naturally occurring human disagreement} at the European Court of Human Rights (ECtHR). We present SV-ECHR, a COC dataset with judge split vote information. We study the disagreement sources among judges from a legal, linguistic, and NLP perspective. We adapt and analyse the task-agnostic taxonomy components of \citealt{xu-etal-2023-dissonance} and introduce SV-specific subcategories. We also quantitatively assess the effects of different subcategories on judges' agreement using proxy variables. The results suggest that disagreements are mainly due to the social-political context of cases.

In addition, we assess the alignment of perceived difficulty between COC models and humans, as well as confidence- and human-calibration. While we see acceptable performance in perceived difficulty and confidence calibration, our analysis indicates suboptimal human calibration. This underscores the necessity for a more in-depth inquiry into methods to better align models’ calibration with human behaviour, highlighting opportunities for further research in this direction.

\section{Related Work}
\paragraph{COC} has been referred to as \textit{Legal Judgment Prediction} (LJP) in previous research \cite{medvedeva2020using,t-y-s-s-etal-2023-zero}. 
There exist numerous works involving corpora from 
ECtHR \cite{aletras2016predicting,chalkidis2019neural,
tyss2023leveraging}. All of these approaches mainly focus on the COC performance of models, which is commonly measured against the human majority class. Our SV-ECtHR dataset extends this work and contains the nuanced judges’ vote information. We proceed to use it to systematically investigate disagreement among judges and the alignment of perceived difficulty between models and judges.
\paragraph{Disagreement / Human Label Variation} is receiving growing attention in mainstream NLP. Various works highlight the presence of HLV, emphasizing the abundance and plausibility of such human disagreements \cite{pavlick-kwiatkowski-2019-inherent, plank-2022-problem}. Researchers advocate for embracing HLV for more trustworthy AI \cite{talat2022machine,casper2023open}. As real-world applications are used to assist diverse audiences, it becomes crucial to investigate and include pluralistic human values in NLP systems \cite{sorensen2023value}. This motivates our study of disagreements among judges' split-votes in the legal decision process. 
Recently, several studies have proposed task-specific taxonomies to identify potential sources of disagreement in various NLP tasks \cite{uma2021learning, sandri-etal-2023-dont, jiang2022investigating}. Building on the meta-analysis of these existing taxonomies, \citealt{xu-etal-2023-dissonance} generalize them to two layers of task-agnostic categories and introduce task-specific categories for COC rationale annotation. 
In this work, we study judges' disagreements in case decision votes, presenting a taxonomy of disagreement with SV-specific subcategories.
\paragraph{Perceived Difficulty} Regarding model difficulty, there is increasing interest in identifying difficult data instances. Various techniques, such as Influence Functions \cite{pmlr-v70-koh17a} and training loss \cite{10.5555/3327757.3327944} have been proposed to identify the difficulty of data instances to a certain model. Pointwise V-usable information (PVI) is a recently introduced difficulty metric \cite{pmlr-v162-ethayarajh22a}, which incorporates mutual information and other types of informativeness\cite{xu2020theory}. 
Despite its recent introduction, PVI has gained significant attention and proven effective in various tasks, including rationale evaluation \cite{prasad-etal-2023-receval} and data selection for augmentation \cite{lin-etal-2023-selective}.  In this work, we leverage PVI to evaluate the alignment of perceived difficulty between human and COC models.
\paragraph{Calibration} Recently in mainstream NLP, researchers posit that the overall reliability of a model is determined by two sides: 1) trustworthiness, which is addressed through models’ confidence measuring, and 2) fairness, which is addressed through their confidence alignment with humans \cite{baan2024interpreting}. Existing calibration studies primarily focus on a classifier's confidence of its predictive performance, commonly evaluated against the human majority class \cite{guo2017calibration, desai-durrett-2020-calibration}. \citealt{baan-etal-2022-stop} instead argue that calibrating against the human majority class is not meaningful in settings with inherent HLV. In our work, we assess the performance and human calibration of COC models in the context of split-vote, a naturally occurring HLV. To the best of our knowledge, this is the first systematic exploration of human calibration in the legal domain.
\paragraph{Uncertainty Evaluation} \label{app:avoidingUE} Within the deep learning community, uncertainty is often classified into two types: aleatoric and epistemic uncertainty. Epistemic (or model) uncertainty arises from a lack of knowledge about the best model, often exacerbated by out-of-distribution examples. Aleatoric uncertainty, on the other hand, stems from inherent ambiguity and can be considered as the variability in experiment outcomes \cite{houlsby2011bayesian,gal2016uncertainty}. As SV is due to inherent disagreement among judges, we regard it as aleatoric uncertainty.
Aleatoric uncertainty is typically quantified through metrics such as Entropy \cite{gal2016uncertainty} and Softmax Response \cite{geifman2017selective}, as recognized in prior works \cite{malinin2018predictive}. However, studies show that these methods are based on predictive entropy, and actually measure total uncertainty, combining both epistemic and aleatoric uncertainty \cite{pmlr-v119-van-amersfoort20a}. Only when we have prior knowledge that either aleatoric or epistemic uncertainty is low, can we use predictive entropy as a suitable measure for the other type \cite{mukhoti2023deep}. Given our small, label-imbalanced, and temporally shifted dataset, we choose to refrain from utilizing these methods to directly measure the aleatoric uncertainty of our cases. Instead, we opt to assess difficulty using PVI, which measures predictive entropy and can be considered an equivalent for total uncertainty.

\section{Legal and Linguistic Backgrounds}

\noindent\textbf{The Legal Decision Making Process} at the ECtHR begins with the applicants lodging their accusation, alleging one or more violations of articles of the European Convention on Human Rights (ECHR). This process aligns with Task B (allegation prediction) in the widely used LexGLUE benchmark \cite{chalkidis2022lexglue}. After receiving the complaint, the court undertakes a review of the case, aiming to determine whether a violation has indeed occurred, corresponding to Task A (violation prediction) in LexGLUE. Cases falling under the purview of well-established ECtHR case-law are directed to a three-judge \textit{Committee}, while others may find themselves before a seven-judge \textit{Chamber}, where decisions are reached through a majority vote. In some circumstances, the \textit{Grand Chamber}, comprising 17 judges, adjudicates cases referred to it by request. 
When rendering a judgment, the Court typically examines only the specific articles alleged by the applicant. This procedure aligns with Task A|B (violation prediction given allegation; \citealt{santosh2022deconfounding}). Our study evaluates the uncertainty of COC models for Task A|B, mirroring the real legal process. 

\begin{figure}[t]
    \resizebox{0.99\linewidth}{!}{
    \centering
    \includegraphics[width = \textwidth]
    {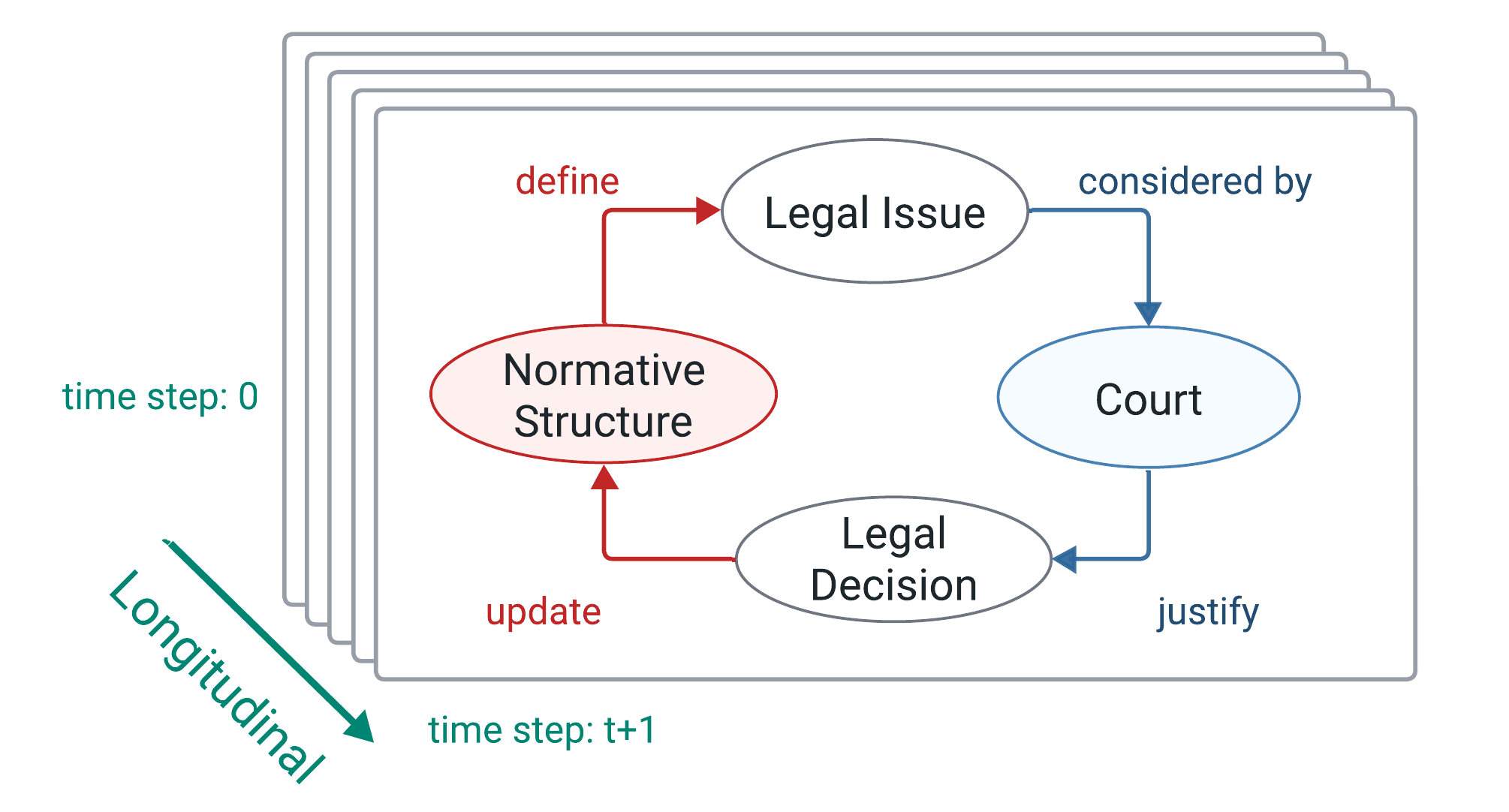}
    }
    \caption{Schematic Diagram of Sources of Disagreement in Legal Decision Process}
    \label{fig:schematic}
\end{figure}

\paragraph{Sources of Disagreement in Legal Decisions} Recent research has delved into sources of HLV and disagreement across various NLP tasks from a linguistic perspective. The classic framework ``Triangle of Reference'' \cite{ogden1923meaning} is widely adopted to categorize disagreement sources in classification tasks \cite{aroyo2015truth, jiang2022investigating}. This concept primarily addresses the relationship between linguistic symbols and the corresponding objects they represent. To study uncertainty sources in NLG tasks, \citealt{baan2023uncertainty} extend it to the ``Double Triangle of Language Production'' catering to the complexities of language generation. However, the legal decision-making process introduces an additional layer of complexity. Legal scholars conceptualize the decision-making process in case-law as circular \cite{ichim2019european}. To better capture this nature, we propose the adoption of the "Direction of Fit" (DoF) framework from speech act theory \cite{searle1985foundations}. In our proposed DoF framework for the legal decision-making process (\autoref{fig:schematic}), judges render decisions based on the interpretation of the existing normative structure of case-law (Law-to-Case). Simultaneously, these case decisions serve as a foundation for future litigation (Case-to-Law). We extend this model with a temporal axis to capture 
the continuous reproduction of the main normative model over time.

\begin{figure*}[t]
    \centering
    \includegraphics[width = \textwidth]
    {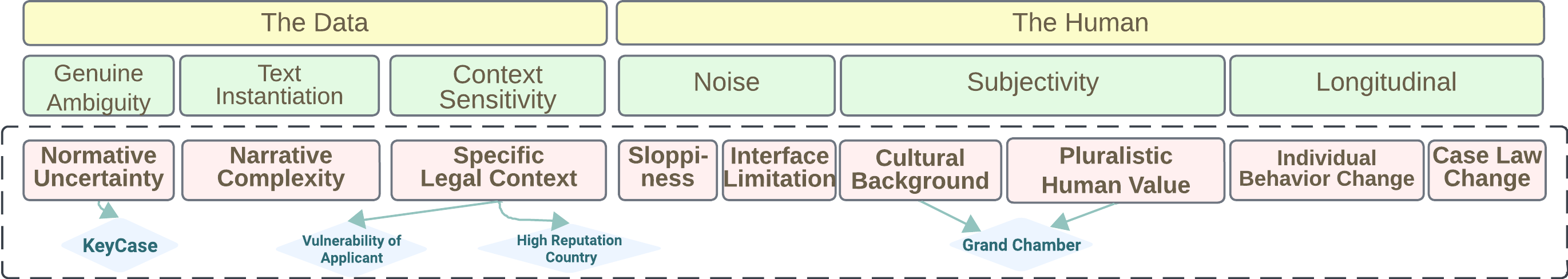}
    \footnotesize\caption{Taxonomy of disagreement sources among judges, with two adapted task-agnostic levels from \citealt{xu-etal-2023-dissonance} and our expanded split-vote specific subcategories and proxy variables (in dashed box).}
    \label{fig:proxy}
    \vspace{-0.2cm}
\end{figure*}

\section{The SV-ECHR Dataset}
\paragraph{Dataset Collection} 
We extract the judge votes distribution for each alleged article\footnote{Following \citealt{chalkidis2022lexglue, santosh2022deconfounding} We use only the 10 most prominent ECHR articles.
} from the public database HUDOC\footnote{https://hudoc.echr.coe.int} using regular expressions. The information about judge votes is always present in the conclusion section
and generally follows certain patterns (See more details in \autoref{app:voteScreenshot}). We did a two-round quality assessment.\footnote{\autoref{app:quailty} offers details on the quality assessment process.} The F1 score of our regular expression rules increased from 0.81 in the first round to 0.98 in the second round. 

\paragraph{Dataset Analysis}\label{sec:data_analysis} 
We then augment the ECHR A|B dataset \cite{santosh2022deconfounding} with our collected vote information according to the document ID. We name the augmented dataset SV-ECHR. It consists of 11k case fact descriptions along with target label information about which convention articles have been alleged to be violated (task B), and which the court has eventually found to be violated (task A), and the judges vote distribution of each alleged article. The dataset is chronologically split into training (2001–2016), validation (2016–2017) and test set (2017-2019) with 9k, 1k, 1k cases respectively. The label set includes 10 prominent ECHR convention \textit{articles}. On average, each \textit{case} has around 1.6 \textit{alleged articles}. Among all 17,604 alleged \textit{case-alleged article pairs} (hereafter \textit{pair}), only around 7\% are split-voted. These numbers underscore the significant label imbalance within the dataset. Additional statistics can be found in \autoref{tab:dataset_stat} in \autoref{app:data_stats}.
To investigate the extent of judges' disagreement in SV-ECHR, we assess the entropies of vote distributions for each \textit{pair}. For the detailed calculation and histogram of the entropy distribution, see \autoref{app:entropy}. We see a large share ($\sim$ 60\% of \textit{chamber} votes) of \textit{single dissenting votes}, where only one single judge voted differently than the six judge majority. A similar pattern is observed for \textit{Grand Chamber} cases involving 17 judges. This pattern aligns with other observations made in empirical legal scholarship (\citealt{fobbe2022introducing}, see \autoref{app:singleJudge}).

\paragraph{Correction of Inconsistent Metadata} \label{sec:meta_correction} \citealt{xu-etal-2023-dissonance} already pointed out inconsistent allegation information in HUDOC. During our quality control process, we found further inconsistent violation information in approximately 2\% of the training set. We updated SV-ECHR with the correct metadata.\footnote{See \autoref{app:meta} for more details of our metadata correction.} Our finding of such inconsistencies calls for mindful data curation when developing COC datasets. 



\section{Disagreement among Judges' Votes}
\subsection{Disagreement Taxonomy}
We use the disagreement taxonomy from \citealt{xu-etal-2023-dissonance} to analyze the reasons behind judges' split votes. \autoref{fig:proxy} displays the taxonomy, with two adapted task-agnostic levels and our expanded split-vote specific subcategories and proxy variables.\footnote{For a comprehensive understanding of each taxonomy category, we direct the reader to \citealt{xu-etal-2023-dissonance}} In the following sections, we provide detailed explanations for each SV-specific category.

\subsubsection{The Data}
Disagreements among judges can stem from various case aspects, including \textit{Genuine Ambiguity} within the Normative Structure, \textit{Narrative Complexity} of the facts, and/or\textit{ Specific Legal Context}. 
\paragraph{Genuine Ambiguity} 
is attributed to \textbf{Normative Uncertainty} of case law in the context of legal NLP (\citealt{xu-etal-2023-dissonance}), which emerges when the court is presented with the possibility of justifying an outcome through multiple legal source interpretations and argumentation. Its occurrence is not uncommon in ECtHR judgments due to the deliberate drafting of the convention in a ‘flexible’ manner to allow tailoring the interpretation to domestic specificities based on the subsidiarity and margin of appreciation principles. It should be noted that there exists an analog \textit{factual} uncertainty where the facts of the case are unclear based on limited (or contradicting) evidence provided. As the ECtHR does not engage in evidentiary reasoning, this aspect is out of scope for this work.


\paragraph{Text Instantiation} covers inconsistency, incompleteness, or biases during text production (i.e. judgment document drafting). When reviewing case files, judges analyze materials that encapsulate the factual background, legal arguments, and evidence presented by both parties. However, these documents are usually produced by the Court Registry, priming the language accordingly. The likelihood of encountering textual unclarity, such as inconsistent framing of facts, increases with the length and complexity of the docket. We refer to this challenge as \textbf{Narrative Complexity}.
\paragraph{Context Sensitivity} 
refers to the \textbf{Specific Legal Context}, characterized by undefined and controversial social-political factors for which judges may not find a precisely tailored legal explanation. We use the term `Context Specificity' because pinpointing the exact arrangement of factors (i.e., attitudinal, normative, and strategic) influencing context in a specific decision is challenging \cite{10.1093/0199256489.003.0001}. Certain social and political factors may only become contentious in a particular context and arrangement of factors.


\subsubsection{The Annotator/Judge} \label{sec:taxonomy_judge}
Disagreements can also arise due to variation in human behavior, which we systematize as \textit{Noise}, \textit{Subjectivity} and \textit{Longitudinal} behavioural change. \textit{Noise} covers errors due to annotator’s \textit{Sloppy Annotation} or \textit{Interface Limitation} \cite{uma2021learning,sandri-etal-2023-dont}. We expect negligence-related noise and interface limitations to be insignificant in naturally occurring ECtHR judge votes.\footnote{Empirical legal scholarship has been investigating the relationship between judge productivity and caseload, including its ramifications on decision quality (e.g., \citealt{engel2020manna}). To the best of our knowledge, no such evidence has been documented in the ECtHR context. 
}


\paragraph{Subjectivity} The ECtHR is composed of judges from the 46 member states 
with diverse legal traditions and cultural backgrounds. Previous work in general NLP demonstrated how annotators’ demographic identities can influence how they label toxicity in text \cite{sap-etal-2022-annotators}. The judges’ \textbf{Cultural Background}, stemming from their legal training and political consideration from their native countries may lead to variations in their rulings compared to those of their colleagues \cite{voeten2008impartiality}.\footnote{See discussion about \textit{National Judge} in \autoref{app:NationalJudge}} Divergent opinions may arise when, for example, two judges from the same country prioritizing different values.  \textbf{Pluralistic Human Values} have recently been made a primary object of mainstream NLP research \cite{sorensen2024roadmap}. Political science research has explored this in the context of  quantitative justice ideal point estimation on the US Supreme Court  (SCOTUS, typically a scalar dimension between liberal and conservative, see \citealt{segal1989ideological}, \citealt{martin2002dynamic}). More similar to our focus on agreement, \citealt{ruger2004supreme} trained an ensemble of court-level and judge-specific decision trees, and observed different performance on conservative and liberal judges.

A comparable liberal vs conservative investigation in the ECtHR context would not be useful. It is composed of 46 Member States, each with its own spectrum of political ideologies, more pluralist and complicated than the liberal-conservative divide. Moreover, there is no temporal stability in the judges’ behavior because they have a limited mandate and act in different bench formations (chamber, grand chamber, etc.). Focusing on individual ideologies to the exclusion of variables accounting for the wider court dynamics could not support accurate analysis.

\paragraph{The Longitudinal Dimension} in \citealt{xu-etal-2023-dissonance} considers two aspects: \textbf{Individual Behavioral Change} over time and \textbf{Case Law Change}. The latter involves shifts in the collective societal attitude towards specific phenomena over time, including changes in laws, policies, and cultural norms. In the legal domain, legal professionals must continually adapt their knowledge, strategies and reasoning to align with the ongoing evolvement of jurisprudence and applicable statutory law. Due to the limited scope of this work, we defer a comprehensive longitudinal study to future research.

\subsection{Conflation of Categories}
It is important to acknowledge that our taxonomy and its categories are not always sharply delineated. Conflation can occur and the boundary between categories may be blurry. Some sources can be interpreted as either Law-to-Case or Case-to-Law. For instance, cases involving vulnerable applicants often involve complex social-political issues lacking clear legal precedent, causing disagreement on the bench. Simultaneously, the judges' subjectivity may lead to varying opinions on whether an applicant should be deemed vulnerable, i.e. judge subjectivity and legal context interact.

\subsection{Proxies \&  Disagreement Correlation}
To quantitatively explore the influence of different taxonomic categories on judges' votes, we work with a legal expert to identify proxy variables. We hypothesize that these proxies correspond to higher disagreement among judges and evaluate the statistical association between them and the entropy of the vote distribution.
If not otherwise mentioned, we retrieve relevant information from the HUDOC dataset and label a case as 1 if it is listed with the proxy feature and 0 otherwise. 

For \textbf{Normative Uncertainty}, we choose \textit{KeyCase} as a proxy 
following \citealt{xu-etal-2023-dissonance}. The court annually chooses a set of ``key cases'', which often deal with complex and novel legal issues. Given the absence of established legal standards for them, they often generate controversy and disagreement among experts. We hypothesize that judges tend to disagree more in ``key cases''.
\paragraph{Specific Legal Context} via two proxy variables: \\ 
\noindent\textit{HighRepCountry} (High Reputation Country): Previous legal and political science work indicates ECtHR judges tend to have more split votes in cases where the defending country is regarded as a High Reputation Country\footnote{Countries with higher democracy index. See \autoref{app:highRepCuntry}.} \cite{dzehtsiarou2015shai}
.\\
\noindent\textit{VulnApplicant} (Vulnerability of Applicant): the ECtHR court adapts convention standards to meet vulnerable individuals’ needs despite  \textit{vulnerability} not being defined in the Convention. It remains underdefined within the context of ECtHR because the judges do not explain the process though which they identify an individual as vulnerable \cite{heriResponsiveHumanRights2021}, which can lead to divergent legal opinions. We retrieve relevant information from the VECHR dataset \cite{xu-etal-2023-vechr} and categorize a case as 1 if its applicant is regarded as vulnerable.\footnote{Only annotation of cases under Article 3 are available in VECHR before 2016. Hence our t-test of \textit{VulnApplicant} is only on cases of Article 3. \autoref{app:vulnerability} offers more explanation on Vulnerability in ECtHR jurisprudence.} 

\paragraph{Subjectivity} refers to the disagreement due to personal opinions and values. We propose that the 17-judge \textit{Grand Chamber} is more prone to disagreement than the 7-judge \textit{Chamber}. This is due to the inherent pluralism of human values in larger groups, which naturally fosters a broader range of viewpoints and thus, increased subjectivity. 
As explained in sec. \ref{sec:taxonomy_judge}, modeling specific ideological dimensions of individual judges is out of scope for this work. We account for some limited political dynamics by including the democratic score of the state against which the claim is brought (the \textit{HighRep} variable above). 

\subsubsection{Do proxy measures correlate with judges' votes?}
To measure the influence of each selected taxonomy category, for each binary proxy variable, we compute the entropy of the vote distribution among all cases exhibiting that variable (`present', value 1) and those that do not (`absent', value 0). We perform an independent t-test to compare the mean entropies between the two groups. 
\begin{table}[th!]
    \centering
        \resizebox{0.46\textwidth}{!}{
\begin{tabular}{|l|rrr|r|}
\hline
 & \multicolumn{2}{l|}{mean entropy} & \multicolumn{1}{l|}{t-value} & \multicolumn{1}{l|}{p-value} \\ \hline
Proxy (absent/present): & \multicolumn{1}{r|}{0} & \multicolumn{1}{r|}{1} & \multicolumn{1}{l|}{} & \multicolumn{1}{l|}{} \\ \hline
GrandChamber & \multicolumn{1}{r|}{0.50} & \multicolumn{1}{r|}{{ 0.49}} & 1.61 & 0.11 \\ \hline
\rowcolor[HTML]{D9EAD3} 
HighRepCountry & \multicolumn{1}{r|}{\cellcolor[HTML]{D9EAD3}0.48} & \multicolumn{1}{r|}{\cellcolor[HTML]{D9EAD3}{ 0.51}} & -4.28 & 2e-5* \\ \hline
\rowcolor[HTML]{D9EAD3} 
 VulnApplicant & \multicolumn{1}{r|}{0.50} & \multicolumn{1}{r|}{{0.62}} & -1.8871 & 0.006* \\ \hline
\rowcolor[HTML]{F4CCCC} 
KeyCase & \multicolumn{1}{r|}{\cellcolor[HTML]{F4CCCC}{ 0.50}} & \multicolumn{1}{r|}{\cellcolor[HTML]{F4CCCC}0.48} & 2.46 & 0.002* \\ \hline
\end{tabular}
}
        \caption{Associations between proxy variables and vote distribution entropy. *: p < 0.05. Green highlighted: confirming our hypothesis. Red highlighted: contradicting our hypothesis.}

        \label{tab:t-test}
\end{table}

\autoref{tab:t-test} shows significantly lower agreement among judges' votes for \textit{HighRepCountry} and \textit{VulnApplicant} (p-value < 0.05). In other words, specific contexts of case facts correlate more with disagreement than Normative Uncertainty. Judges express disagreement related to how facts fit the norm because of circumstances related to the social standing of the applicant and the political situation of the defending State. This is an inherent characteristic of adjudication where judges qualify the context so as to fit the norm. The inclusion of a category related to the legal norm's scope could have provided further insights. 

Our hypothesis regarding \textit{GrandChamber} does not hold precisely because it does not matter which bench formation deals with the case - Chamber or Grand Chamber. What matters is how judges decide to fit the facts to the norm.

Interestingly, we found less disagreement among judges on \textit{KeyCase}. Experts suggest this may stem from the Court's assigning the importance score ex post, after interpretation and agreement are reached. The Court is more likely to designate a consensus case as \textit{KeyCase} to maintain coherent jurisprudence, rather than a controversial case that could invite future applicants to base cases on disputed facts and dissenting opinions. As reflected in our DoF framework (\autoref{app:dof}), Case-to-Law uncertainty inherently perpetuates Law-to-Case uncertainty. 

    \begin{table*}[th!]
          \centering
    \resizebox{\linewidth}{!}{
\begin{tabular}{l|l|ccc|ccc|ccc|c}
\hline
\multicolumn{1}{c|}{} &  & \multicolumn{3}{c|}{hm-F1 $\uparrow$} & \multicolumn{3}{c|}{ECE $\downarrow$} & \multicolumn{3}{c|}{DistCE $\downarrow$} & count \\ \hline
\multicolumn{1}{c|}{} &  & \textbf{/} & TS & soft & / & TS & soft & / & TS & soft & \multicolumn{1}{l}{} \\ \hline
\multicolumn{1}{c|}{LegalBERT} & u & \textbf{69.30 ± 1.88} & \textbf{69.30 ± 1.88} & 67.31 ± 0.5 & 23.32 ± 1.01 & \textbf{2.95 ± 0.67} & {\ul 22.04 ± 0.63} & {\ul 25.10 ± 1.17} & 37.23 ± 1.71 & \textbf{24.70 ± 0.17} & 1463 \\
\multicolumn{1}{c|}{} & sv & \textbf{53.67 ± 4.59} & \textbf{53.67 ± 4.59} & 46.21 ± 4.41 & 29.92 ± 2.81 & \textbf{8.02 ± 0.58} & {\ul 28.49 ± 0.82} & 41.03 ± 2.75 & \textbf{28.28 ± 0.99} & {\ul 40.72 ± 1.00} & 112 \\
 & all & \textbf{68.03 ± 1.29} & \textbf{68.03 ± 1.29} & 65.16 ± 0.82 & 23.75 ± 1.08 & \textbf{2.99 ± 0.83} & {\ul 22.32 ± 0.64} & {\ul 26.23 ± 1.0} & 36.60 ± 1.52 & \textbf{25.84 ± 0.21} & 1575 \\ \hline
\end{tabular}

}

        \footnotesize\caption{COC performance (hm-F1), confidence-calibration (ECE), and human-calibration (DistCE) performance with std ($\pm$) on test set. "/": COC finetuned models in \autoref{sec:coc}; "soft": models fine-tuned with soft-loss; "TS": models after Temperature Scaling.
        Results shown over 3 random seeds. See \autoref{tab:calibration_results_app} in \autoref{app:moreHumanCali} for performance on dev set.}
    \label{tab:calibration_results}
    \end{table*}

    \begin{figure*}[h]
    \centering
    \includegraphics[width = \textwidth]{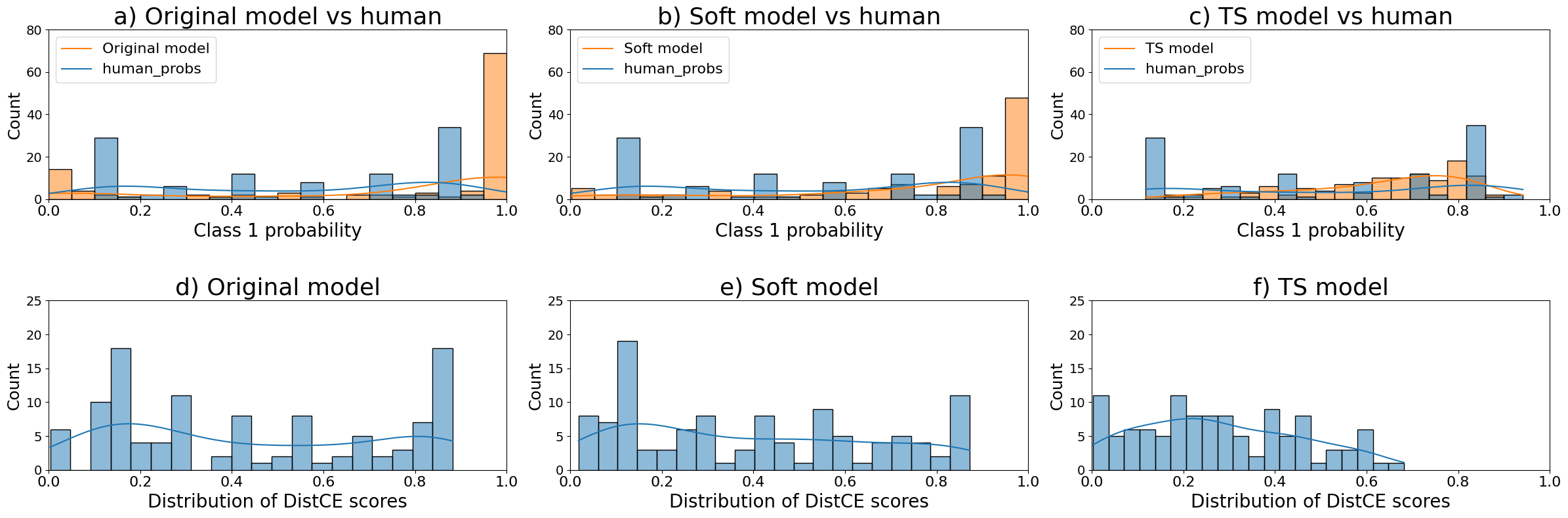}
    \caption{
    The distributions over probabilities for class 1 of the models vs human vote distributions (row 1) and distCE (row 2). See \autoref{fig:appCali} in \autoref{app:moreHumanCali} for more figures comparing human uncertainty to model uncertainty}
    \label{fig:cali_resul_l-1_test}
\end{figure*}

\section{Judge-Model Misalignment}
\vspace{-0.08cm}
\subsection{Experimental Setup of COC} \label{sec:coc}

We assess the alignment of a model's prediction difficulty\footnote{Another approach would be assessing a model's prediction \textit{uncertainty}. Refer to \autoref{app:avoidingUE} for an explanation of our decision to abstain from using Uncertainty evaluation methods} with judge disagreement, specifically COC models trained on ECtHR Task A|B. 
To address input length constraints, we employ a LegalBERT \cite{chalkidis2020legal} variant of the hierarchical attention model\footnote{See \autoref{app:arch} for model structure and implementation.} \cite{yang2016hierarchical}, as adapted from \citealt{santosh2022deconfounding}.
\paragraph{Task} Benchmark task ECtHR A|B (Violation Identification given Allegation) identifies the set of violated convention article(s) from the textual case facts and list of allegedly violated articles. 
\paragraph{Models} \citealt{santosh2022deconfounding} encodes allegation information as a multi-hot vector concatenated with the text representation. By contrast, in this study, we textify allegation information as “Alleged Article: X”, where is X is a comma-separated list of articles and prepend to the input text. We employ the LegalBERT (\texttt{nlpaueb/legal-bert-base-uncased}; \citealt{chalkidis2020legal}) as the backbone in hierarchical models.\footnote{To explore the impact of domain-specific pre-training, we did also experiment on BERT. See \autoref{app:bert}.}
\paragraph{Metrics} Our evaluation of COC performance employs the micro-F1 (mic-F1), macro-F1 (mac-F1), and hard-macro-F1 (hm-F1;~\citealt{santosh2022deconfounding}). The hm-F1 is calculated as the mean F1-score for each article, where cases with that article having been violated are considered as positive instances, and cases with that article being alleged but not found to be violated as negative instances, resulting in a smaller pool of more difficult negatives. 
\paragraph{COC Performance}\label{sec:cocPerformance}  
We run experiments across three random seeds. The hierarchical LegalBERT achieves 73.55 ± 0.62 for mic-F1, 68.75 ± 1.13 for mac-F1 and 68.03 ± 1.29 for hm-F1. Although the mic-F1 and mac-F1 scores closely align with those reported in \cite{santosh2022deconfounding,xu-etal-2023-dissonance}, differences of $\sim$ 2\%, the hm-F1 performance demonstrate a noteworthy improvement of $\sim$10\%.\footnote{This can be partly attributed to our correction of labels (see \autoref{sec:meta_correction}). Some variation may also be due to randomness in weight initialization. See our discussion in \autoref{app:ablation}} Importantly, hm-F1 exhibits particular sensitivity to allegation information. For the rest of this paper, we adopt hm-F1 as our primary metric for COC performance.

    

\subsection{Difficulty of SV Cases in COC} 

Pointwise V-usable information (PVI;~\citealp{pmlr-v162-ethayarajh22a}) measures the difficulty of an instance within a dataset for a given model as: $PVI(x\rightarrow y) = -\log_2 g[x](y) + \log_2 g'[\emptyset](y)$. 
Computing PVI involves fine-tuning a model $g$ on two datasets. The first dataset $D$ comprises input-target pairs $\{(x_i, y_i) | (x_i, y_i)\in D\}$, while the second $D'$  is used to fine-tune model $g'$ on null-target pairs, $\{(\emptyset, y_i) | (x_i, y_i)\in D\}$ ($\emptyset$ represents an empty string intended to fit the label distribution). PVI serves as the measure of information gain resulting from the provision of an input during fine-tuning. Higher PVI suggests a better representation of the input in the model, and thus an easier instance.

 \begin{table}[th!]
    \centering
     \resizebox{\linewidth}{!}{
    \begin{tabular}{| l  |l  |l  |l  |l  |l  |l |}
    \hline
     & \multicolumn{2}{|l|}{hm-F1} & \multicolumn{2}{|l|}{mean PVI} & t-value & p-value \\
    \hline
     & U. & SV & U. & SV &  &  \\
    \hline
    LegalBERT& 69.3 ± 1.88 & 53.67 ± 4.59 & 0.77 & -0.12 & 2.66 & 0.008 \\
    \hline
    
    \end{tabular}
    }
    \footnotesize\caption{COC prediction (hm-F1) and difficulty scores (mean PVI) for unanimous (U.) and SV cases. } 
    \label{tab:pvi_groups}
            \vspace{-0.1cm}
\end{table}
\vspace{-0.1cm}

\paragraph{Experiment Settings} We use the COC fine-tuned models from Sec \ref{sec:coc} as the input-target model $g$. We fine-tune an architecturally identical model $g'$ with the input replaced by $\emptyset$. Subsequently, we partition the test set into two distinct partitions of SV and unanimous cases, respectively. Our expectation is that the SV subset will exhibit lower average PVI compared to the unanimous subset. We use entropy of the judges' vote distribution as an estimator of case difficulty: the more the judges disagree with each other, the harder a case is. 
\paragraph{Results} \autoref{tab:pvi_groups} presents the average COC performance (hm-F1) and model-perceived difficulty (mean PVI scores). The model exhibits lower hm-F1 and PVI scores for SV cases compared to unanimous cases. An independent t-test on the average PVI scores between the two groups reveals a significant difference ($p<0.05$). This indicates that SV cases are more challenging for the models than unanimous cases, aligning with human perceptions of difficulty. We also calculate the Pearson correlation coefficients between PVI scores and the entropy of judges’ vote distribution of SV cases. 
LegalBERT has a  correlation coefficient (r-value) of -0.068 and a p-value of 0.48. The results reveal a negative correlation between PVI scores and the degree of disagreement among judges, consistent with our expectations. However, the observed correlation is very weak ($|r|<0.1$) and statistically insignificant ($p >0.05$). This suggests that, while the model can capture differences in difficulty between unanimous and SV cases, it struggle to accurately represent the nuanced degree of judges' disagreement. Therefore, we next evaluate the models' confidence and soft-label training.

\subsection{Calibration of SV Cases}

\subsubsection{Evaluation Metrics}


Most current NLP research focuses on \noindent\textbf{Confidence Calibration} \cite{jiang-etal-2021-know, desai-durrett-2020-calibration}: \textit{The model should be unsure when it does not know the answer}. A model is considered well confidence-calibrated if its prediction confidence aligns with its predictive accuracy, commonly evaluated against the human majority class. Yet there is a growing interest in accounting for HLV and/or pluralistic values \cite{plank-2022-problem, sorensen2023value}. Therefore we extend our evaluation to include \noindent\textbf{Human Calibration} \cite{baan-etal-2022-stop}: \textit{The model should be unsure when humans are unsure about the answer}. We consider a model well human-calibrated if the categorical distribution of predicted class probabilities align well with the actual human vote distribution.

\subsubsection{Calibration Methods}
For \textit{confidence calibration}, we employ \textbf{Temperature Scaling} (TS;~\citealt{guo2017calibration}). This simple yet widely used post-hoc method uses a single temperature parameter $t$ to scale the output logits of a classifier. We choose the temperature $t$ by searching a range of possible values for $t$ on the dev set. For\textit{ human calibration}, we adopt the approach from \citealt{Peterson_2019_ICCV, uma2020case} by finetuning with a \textbf{Soft Loss Function}.\footnote{See \autoref{app:soft} for details of the Soft Loss Function.} During training, models are exposed to \enquote{soft labels} derived from the judges' vote distributions, serving as target distributions in a cross-entropy loss function. 
\paragraph{Expected Calibration Error} (ECE) is the most often used metric for \textit{confidence calibration} \cite{naeini2015obtaining, guo2017calibration}. A lower ECE indicates better calibration, suggesting that the model's predicted probabilities are more accurate reflections of the true probabilities. Refer to the \autoref{app:ece} for further details on ECE.

\paragraph{DistCE} by \citealt{baan-etal-2022-stop} measures \textit{ human calibration}, which can be calculated as $\mathrm{DistCE}(x)=1 / 2\|\mathbf{q}-\mathbf{p}\|$, where $q$ is the vote distribution and $p$ is the model's predictive distribution.






\subsubsection{Results \& Discussion}
\autoref{tab:calibration_results} presents the experimental results. We make the following observations:
\paragraph{i) SV cases are indeed the most challenging.} It is reflected by their lowest COC performance (hm-F1). They also exhibit a higher degree of miscalibration (higher ECE) and misalignment with human responses (higher distCE). 
\paragraph{ii) Applying TS greatly enhanced confidence calibration.} \autoref{tab:calibration_results} presents that applying TS has greatly reduced the ECE score, without negatively impacting COC performance (hm-F1).
\paragraph{iii) The models remain misaligned with the human vote distribution.}
\autoref{tab:calibration_results} shows that soft-loss tuning only slightly improves the alignment between the model's prediction confidence and the human vote distribution (lower DistCE). \autoref{fig:cali_resul_l-1_test} illustrates that Soft LegalBERT exhibits noticeably fewer instances of over-confident predictions when judges do not unanimously agree, as depicted on near-0/1 probability portions of \autoref{fig:cali_resul_l-1_test}b, in contrast to the original model shown in \autoref{fig:cali_resul_l-1_test}a. 
On split-vote cases,  TS models exhibit substantially lower DistCE scores than soft models. However, we do not consider that TS provides a better human calibration. The lower DistCE scores may be attributed to an overly aggressive temperature, resulting in a more uniform output distribution (see \autoref{fig:cali_resul_l-1_test}c, with a temperature of 5.5). Moreover, judge votes often exhibit a quasi-bimodal distribution, with many split votes caused by a single judge dissenting pro/con the finding of a violation (see discussion on \textit{single dissenting votes} in \autoref{sec:data_analysis}). Further, the DistCE score histograms (\autoref{fig:cali_resul_l-1_test}d-f) illustrate that soft-tuning (\autoref{fig:cali_resul_l-1_test}e) improves human alignment by reducing instances of extreme miscalibrations in the right tail, as compared to the original model in \autoref{fig:cali_resul_l-1_test}d. It is noteworthy that the TS model (\autoref{fig:cali_resul_l-1_test}f) shifts the distribution further to the left by reducing the right tail. However, this is at the expense of predictions that perfectly align with human judgement probabilities, as evidenced by fewer instances of DistCE scores below 0.15 in the TS model compared to the Soft model. Therefore, despite TS models potentially displaying lower DistCE scores than soft models,  they do not provide an optimal fit. It is crucial to analyze the distributions prior to drawing conclusions, especially as we have shown in cases with split votes, where distributions are \textit{bimodal}.

\section{Conclusion}

We present \textsc{SV-ECHR}, a new COC dataset enriched with naturally occurring split-votes by judges on the ECtHR. We also present a SV-enriched taxonomy of disagreement sources. Our experiments reveal shortcomings when combining TS and ECE to improve and measure calibration of models against labels subject to inherent human disagreement. This is due to a distribution mismatch reflected in low DistCE scores. Soft loss training produces only slightly better human calibration scores.
We call on the community to explore methods for measuring and improving the alignment of model calibration with human behavior; as well as more research into incorporating HLV in NLP. 

\section*{Limitations}
Our study is constrained by the datasets, models, and selective prediction techniques under consideration, primarily relying on the ECtHR dataset. Expanding the investigation to encompass diverse datasets and legal jurisdictions would enhance our understanding of disagreement in judge decision votes and the alignment of perceived difficulty between judges and models.

Additionally, due to computational limitations, we are constrained from pre-training language models from scratch or fine-tuning large language models (LLMs). Our study relies on existing pre-trained BERT-based models, focusing solely on fine-tuning. We refrain from exploring LLM models, as no widely agreed-upon method for measuring calibration for LLMs has emerged at the time of submission. Furthermore, with respect to variations introduced by prompts and data contamination during pretraining, exploring the use of LLMs for difficulty perception and calibration on a small-scale, specialized legal dataset is a distinct research question deserving a separate paper.

\section*{Ethics Statement}
In this study, our retrieved judges' vote information from the publicly available HUDOC dataset, with the overarching goal of improving the alignment of perceived difficulty and calibration between models and judges. While these decisions include real names and are not anonymized, we do not anticipate any harm beyond the availability of this information resulting from our experiments.

The task of case outcome classification raises significant ethical and legal concerns, both generally and specifically concerning the European Court of Human Rights \cite{medvedeva2020using}. It is important to clarify that we do not advocate for the practical implementation of COC within courts. Previous work \cite{santosh2022deconfounding} has demonstrated that these systems heavily rely on superficial and statistically predictive signals lacking legal relevance. This underscores the potential risks associated with employing predictive systems in critical domains like law and highlights the importance of trustworthy and explainable legal NLP. 

In this work, we investigate the sources of disagreements among judges in their decisions on case facts. While technically situated outcome classification models, we intend our analysis of different types of disagreement to promote the acceptance of human label variation and pluralistic human values within the legal NLP community. By acknowledging and understanding various perspectives, interpretations, and biases of judges, we contribute to a more comprehensive and inclusive discourse within the field. A continuation of this work can unfold long term practical implications: By identifying patterns which create uncertainty, applicants could potentially ‘exploit’ the distinctive circumstances that change the normative assessment for the purpose of their own case, or otherwise inform litigation strategy. By the same token, however, it would offer the court the possibility to investigate and ‘check’ whether it applies legal norms coherently, in line with the demands of consistency, foreseeability and certainty. The potential effects of data-driven decision making in the legal domain cut both ways, and must be reconciled mindfully.

\section*{Acknowledgements}
We would like to thank the anonymous reviewers for their valuable comments and suggestions. We would also like to thank the members of the MaiNLP/CIS research lab for their thoughtful feedback. We acknowledge the support of our funders. BP is supported by the ERC Consolidator Grant No.\ 101043235.
\bibliography{custom}

\appendix


    \begin{figure*}[th!]
        \centering
        \includegraphics[width = 0.98\textwidth]
        {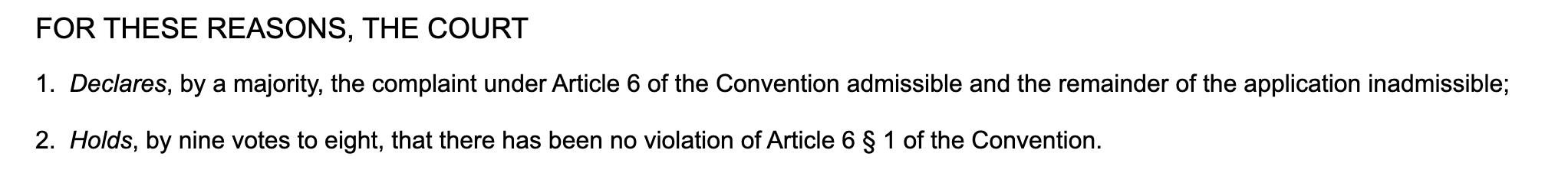}
        
        \caption{Screenshot of Judges' vote information in HUDOC}
        \label{fig:voteScreenshot}
    \end{figure*}

\section{Information about Judge' Votes in HUDOC}\label{app:voteScreenshot}
We observe that the information about judge' votes is always present in conclusion section of the decision towards the end and generally following certain patterns, such as: "Holds, by X votes to Y, that there has been a/no violation of Article Z of the Convention"; or "Holds by X votes to Y that Article Z has (not) been violated" as in \autoref{fig:voteScreenshot}.

\section{Quality Validation of SV-ECHR} \label{app:quailty}
To ensure the quality of the dataset, we did a two-round quality assessment. In the first round,  we manually retrieved the split-vote information of 20 cases as gold labels. We then assess the votes information exacted by our regex rule with the gold labels. We analysis the error and improve our regex rule. In the second round, we repeat this process with another 20 cases. The F1 score increased from 0.81 in the first round to 0.98 in the second round.

\section{Dataset Statistics} \label{app:data_stats}
\autoref{tab:dataset_stat} offers additional statistics on the SV-ECHR dataset.

\begin{table}[h]
    \resizebox{0.99\linewidth}{!}{
\begin{tabular}{|l|c|c|c|c|}
\hline
 & \# Cases & \begin{tabular}[c]{@{}c@{}}\# Case\\ -Article Pairs\end{tabular} & \begin{tabular}[c]{@{}c@{}}\# alleged Case-\\ Article Pairs\end{tabular} & \begin{tabular}[c]{@{}c@{}}\# SV Case-\\ Article Pairs\end{tabular} \\ \hline
Train & 9000 & 90000 & 14513 & 960 \\ \hline
Dev & 1000 & 10000 & 1516 & 135 \\ \hline
Test & 1000 & 10000 & 1575 & 112 \\ \hline
\end{tabular}

}
    
    \caption{More statistics of the SV-ECHR dataset. }
    \label{tab:dataset_stat}
\end{table}

\section{Correction of Inconsistent Metadata} \label{app:meta}

\begin{table}[h]
        \resizebox{0.95\linewidth}{!}{
\begin{tabular}{|l|l|l|l|l|}
\hline
 & \# Case-Article Pairs & \begin{tabular}[c]{@{}l@{}}\#  alleged \\ pairs\end{tabular} & \begin{tabular}[c]{@{}l@{}}\# pair w/. \\ wrong alleg.\end{tabular} & \begin{tabular}[c]{@{}l@{}}\# pair w/. \\ wrong vio.\end{tabular} \\ \hline
Train & 90000 & 14513 & 2719 & 292 \\ \hline
Dev & 10000 & 1516 & 247 & 0 \\ \hline
Test & 10000 & 1575 & 278 & 0 \\ \hline
\end{tabular}
}
    \caption{Statistics about the corrected meta information. }
    \label{tab:meta}
\end{table}

\autoref{tab:meta} shows the statistics about our correction of case metadata. Notably, apart from the inconsistent allegation information pointed out by \citealp{xu-etal-2023-dissonance}. We further found approximately 2\% inconsistent violation information in the train set during our quality control process. Most of the cases are so-called `striked-out' cases which are difficult to parse. We also find finetuning on the corrected train set improves model's COC performance (\autoref{sec:cocPerformance}) as discussed in \autoref{app:ablation}. 

    

\section{Entropy of Judges' Vote Distribution} \label{app:entropy}

\begin{figure}[]
    \centering
    \includegraphics[width = 0.48\textwidth]
    {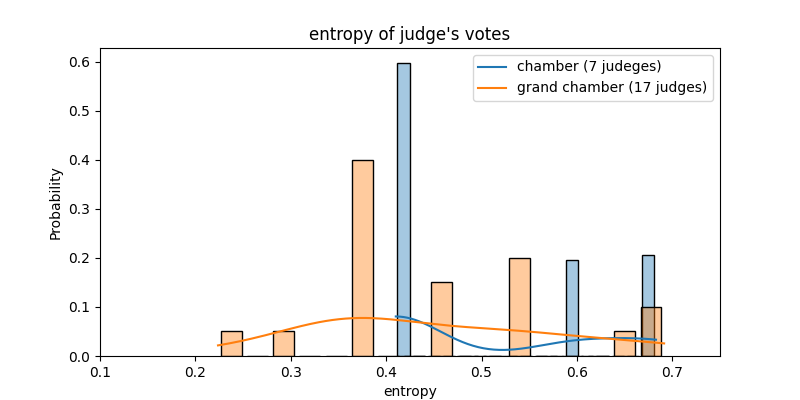}
        \caption{Histogram of entropy of judges' votes distribution over train/dev/test split}
    \label{fig:entropy_chamber}
\end{figure}

\autoref{fig:entropy_chamber} display the histogram of the entropy of each pair's vote distribution. 
The entropy is calculated as $\mathbf{H}(\mathbf{p})=-\sum_{i \in \mathcal{C}} p_i \log \left(p_i\right)$ and $p_i=\frac{n_i}{\sum_{j \in \mathcal{C}} n_j}$ , where $\mathcal{C}$ is the label category set [violation, non-violation] and $n_i$ is the number of judges voting for category $i$.

\section{Debates in Legal Scholarship about Single Judge Dissenting Vote} 
\label{app:singleJudge}

\autoref{tab:dataset_stat}  reveals that $\sim$ 60\% of \textit{chamber} votes (involving 7 judges) exhibit entropy of around 0.4, indicating that only one judge vote differently from the remaining 6 judges (entropy([1,6]) $\approx$ 0.4). A similar pattern is observed for\textit{ Grand Chamber} cases involving 17 judges. This \textit{Single Judge Dissenting} pattern aligns with ongoing debates in legal scholarship. For instance, \citealt{fobbe2022introducing} examined the number of dissenting opinions in decisions from the International Court of Justice (ICJ). Their results indicate a significant proportion of unanimous decisions, followed by a monotonously decreasing number of dissents.
Some legal scholars support the view that diverging opinions do not signal a division of the bench inside the Court over the scope of protection of rights, even less a lowering of standards of protection. Others support the contrary view according to which dissenting opinions send a signal of walking-back in terms of effective protection \cite{helfer2020walking}. \autoref{tab:dataset_stat} and \citealt{fobbe2022introducing}'s finding actually shows that the case law does not consistently, but rather seldomly, give rise to dissenting opinions. Moreover, one can not ignore the national judge's dissenting vote (\autoref{app:NationalJudge}). It is difficult to assess the weight of a single judge dissenting on the decision of the court as a whole. Often times, dissenting opinions are not followed up on. Our focus here is limited to developing a yardstick against disagreement in human decisions can be measured.

\section{Judge's Negligence Errors} \label{app:hungry}
Empirical legal scholarship identified time constraints leading to judge negligence errors in certain jurisdictions like the US Board of Veterans’ Appeals \cite{ho2019quality} and Israeli Magistrate Courts \cite{engel2020manna}. However, to the best our knowledge, there is no evidence has been documented in the ECtHR context for  careless behavior by judges. The infamous Hungry Judge Effect, stemming from work in \cite{danziger2011extraneous}, suggested that judges tend to issue harsher sentences just before lunch, presumably influenced by hunger. 
However, a subsequent study \cite{weinshall2011overlooked} argues that the observed peak in favorable decisions after a meal break is likely an artifact of case presentation order, considering anticipated outcomes and duration. Overall, this rebuts the notion that hunger impacts judges' rational decision-making.

\section{National Judge} \label{app:NationalJudge}
In the ECtHR context, a \textit{national judge} refers to the judge appointed from the respondent country of the case which was brought before the court. Like all other judges, national judges maintain independence and do not act as representatives of their respective governments. However, as previously noted in studies by legal schorlars, national judges have been observed to dissent more frequently in cases finding a violation of the Convention compared to their non-national counterparts \cite{kuijer_1997, helfer2020walking}.\citealt{voeten2008impartiality} also provides evidence and explanations, including for why the judges are considered policy-seekers, and concludes  that judicial activism is driven by the political logic of European integration.

\section{Key Case} \label{app:keycase}
\noindent\textit{KeyCase}: The ECtHR annually chooses a set of significant cases, known as ``key cases''. These often deal with complex and novel legal issues. Given the absence of established legal standards for interpreting them, they often generate controversy and disagreement among judges.


\section{High Reputation Country} \label{app:highRepCuntry}
Previous legal and political science work indicates ECtHR judges tend to have more split vote in cases where the defending country is regarded as a `High Reputation Country'. Strong democracies enjoy a high reputation in front of the judges while weak and new democracies only benefit from a low reputation implying that the ECtHR should issue more demanding judgments against low reputation states than against high reputation states, with the view not to damage its own reputation and ensure compliance with its decisions \cite{dzehtsiarou2015shai}.
We extracted the country information from the HUDOC metadata, namely Respondent State(s). Following \citealt{chalkidis-etal-2022-fairlex}, we group the countries according to the disproportion of violations between eastern and central European countries, and the rest of European countries (western European, Nordic, mediterranean states).

\section{Vulnerability in ECtHR} \label{app:vulnerability}
The ECtHR adapts convention standards to meet individual needs and to ensure effective human rights protection. Recognizing vulnerability is crucial for understanding unique needs and implementing targeted support systems. However, the concept of `vulnerability' remains undefined by the court, which can lead to divergent opinions on what qualifies as vulnerability from a legal point of view. Cases involving vulnerable applicants often deal with complex social-political issues related to the protection and rights of individuals who may face various challenges, such as victimization, migration, discrimination, reproductive health, unpopular views etc.
The inclusion of vulnerability as a proxy variable is an example of how a human rights legal concept that is difficult to define can potentially give rise to diverging opinions on the bench. A comprehensive study of such concepts in ECtHR jurisprudence lies beyond the scope of this work.
We direct the reader to \citealt{heriResponsiveHumanRights2021} for a systematic legal study and \citealt{xu-etal-2023-vechr} for comprehensive NLP research on vulnerability in the ECtHR context.







\section{The impact of domain-specific pre-training on
uncertainty representations} \label{app:bert}

\autoref{tab:coc_performance_app} reports the results of classification performance. Notably, the model with legal-specific pre-training (LegalBERT) outperforms the one with general pre-training (BERT). 

\autoref{tab:calibration_results_app} shows that fine-tuning with soft-loss to human labels yields minimal ECE changes with a discrepancy: BERT shows a slight decrease, while LegalBERT displays a minor increase. This mirrors the issue identified by \citealt{baan-etal-2022-stop}, highlighting the challenge of using ECE to measure calibration on disagreement data. They argue that even a classifier perfectly modeling human judgment distribution would still be severely miscalibrated when measured by ECE.

   \begin{table}[h!]
        \centering
        \resizebox{0.9\linewidth}{!}{
        \begin{tabular}{| l | l | l | l |}
        \hline
         & mic-F1 & mac-F1 & hm-F1 \\
        \hline
        BERT& 72.31 ± 4.09 & 65.53 ± 6.22 & 64.22 ± 3.46 \\
        \hline
        LegalBERT& 73.55 ± 0.62 & 68.75 ± 1.13 & 68.03 ± 1.29 \\
        \hline
        
        \end{tabular}
        }
        \footnotesize\caption{COC performance on test set}
        \label{tab:coc_performance_app}
    \end{table}

\section{Ablation Experiment} \label{app:ablation}

    \begin{table}[]
            \resizebox{0.95\linewidth}{!}{
\begin{tabular}{|l|l|l|l|}
\hline
finetune dataset & mic-F1 & mac-F1 & hm-F1 \\ \hline
ours & \textbf{73.55 ± 0.62} & \textbf{68.75 ± 1.13} & \textbf{68.03 ± 1.29} \\ \hline
\citealt{santosh2022deconfounding} & 73.41 ±  2.5 & 67.74± 3.2 & 63.93±  1.7 \\ \hline
\end{tabular}
}
        \caption{LegalBERT's COC performance on test set with different finetune dataset. \textit{Our} dataset with correction of metadata as mentioned in \autoref{sec:meta_correction}}
        \label{tab:ablation}
    \end{table}

We also train hierachical LegalBERTs with the textified allegation information on the original dataset from \cite{santosh2022deconfounding} without corrected metadata. Results in \autoref{tab:ablation} show that model achieves better performance when fine-tuned on our dataset with corrected metadata.

\begin{figure}[]
    \centering
    \includegraphics[width = 0.25\textwidth]
    {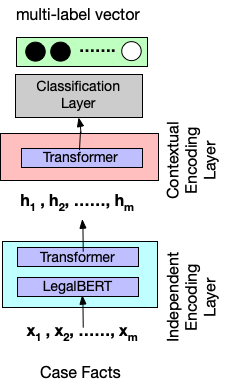}
    \caption{Hierarchical Classification Variant} \label{fig:arch_a}

\end{figure}

\section{Details of the Models} \label{app:arch}
\subsection{Architecture of the Hierarchical Model} 
For the \emph{hierarchical} variant of pre-trained BERT models, we use a greedy input packing strategy where we merge multiple paragraphs 
into one packet until it reaches the maximum of 512 tokens. We independently encode each packet of the input text using the pretrained model and obtain representations for each packet. Then we apply a non-pretrained transformer encoder to make the packet representations context-aware. \autoref{fig:arch_a} illustrates the detailed architecture of the hierarchical model.

\subsection{Implementation Details}
We use \emph{BERT} "bert-base-uncased" \cite{kenton2019bert}, and \emph{LegalBERT} "nlpaueb/legal-bert-base-uncased" \cite{chalkidis2020legal} from the Transformers Hub \cite{wolf-etal-2020-transformers} as our backbone models.\\
\textbf{Hyperparameter \& Overfitting Measures}: 
For the hierarchical models, we employ a maximum sentence length of 128 and document length (number of sentences) of 80. The dropout rate in all layers is 0.1. We follow the hyperparameters from \citealt{chalkidis2022lexglue} with a batch size of 8 and learning\_rate of 3e-3. We train models with the Adam optimizer for up to 10 epochs. We use \emph{PyTorch} \cite{NEURIPS2019_9015} 2.0.1.

\section{Soft Loss Function} \label{app:soft}
The formulation of the soft loss function is represented as \\$-\sum_{i=1}^n \sum_c p_{h u m}\left(y_i \mid x_i\right) \log p_\theta\left(y_i=c \mid x_i\right)$, 
where we compute $p_{\text {hum }}$ by applying a standard normalization function to the judges' votes for each pair following \citealt{Peterson_2019_ICCV}. 

\begin{figure*}[t]
    \centering

        \includegraphics[width=1\linewidth]{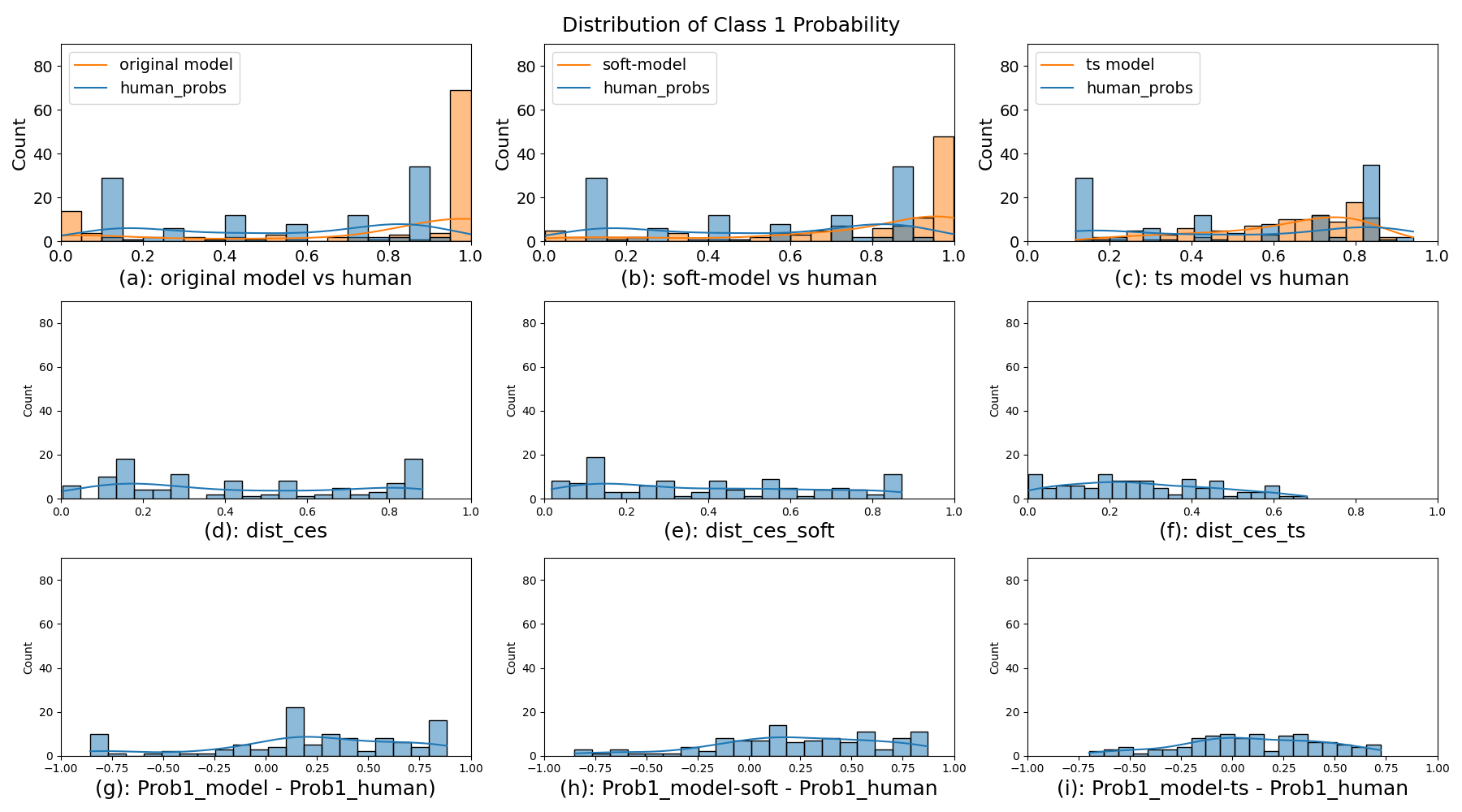}
    
    \caption{LegalBERT seed 1 on the SV cases in test set. Row 1: the distribution over instance-based absolute errors between probabilities for class 1 of the model vs human vote distributions. Row 2: the distribution over instance-based DistCE of mdodels. Row 3: the distribution over errors between probabilities for class 1 of the model vs human vote distributions.}
    \label{fig:appCali}
\end{figure*}

\section{Expected Calibration Error}\label{app:ece}
Expected Calibration Error (ECE) is a measure of the difference between the predicted probabilities assigned by a model and the accuracy of those predictions. ECE is typically calculated by dividing the predicted probability space into a fixed number $M$ of intervals (or bins) and then computing the average absolute difference between the predicted and observed probabilities within each bin $B_m$ as follows:
\begin{equation}
\mathrm{ECE}=\sum_{m=1}^M \frac{\left|B_m\right|}{N}\left|\operatorname{acc}\left(B_m\right)-\operatorname{conf}\left(B_m\right)\right|
\end{equation}

A lower ECE indicates better calibration, suggesting that the model's predicted probabilities are more accurate reflections of the true likelihood.

\section{More Calibration Results} \label{app:moreHumanCali}

\autoref{tab:calibration_results_app} presents the calibration experiment results on both dev and test set. Soft-training  improves the model's COC performance (hm-F1) on SV cases for the development set, but lowers it for the test set. Two contributing factors account for this discrepancy: i) SV instances constitute less than 10\% of the total dataset, potentially limiting the generalizability of improvements. ii) Given the temporal split of the dataset, distribution shifts in case-law may occur. For instance, legal issues addressed in SV cases could stabilize after a key legal decision. In other words, something that is legally controversial and uncertain in the training set time period may be settled law in the development and test dataset partitions. Our models may have overfit to the training set period.
\autoref{fig:appCali} shows more figures comparing human uncertainty to model uncertainty.

    \begin{table*}[th!]
         \centering
    \resizebox{0.95\linewidth}{!}{

\useunder{\uline}{\ul}{}
\begin{tabular}{l|l|ccc|ccc|ccc|c}
\hline
 &  & \multicolumn{3}{c|}{{\color[HTML]{3B3B3B} hm-F1}} & \multicolumn{3}{c|}{{\color[HTML]{3B3B3B} ECE}} & \multicolumn{3}{c|}{{\color[HTML]{3B3B3B} DistCE}} & {\color[HTML]{3B3B3B} count} \\ \hline
 &  & / & TS & soft & / & TS & soft & / & TS & soft & \multicolumn{1}{l}{} \\ \hline
BERT & {\color[HTML]{3B3B3B} u} & {\color[HTML]{3B3B3B} \textbf{66.19 ± 3.55}} & {\color[HTML]{3B3B3B} \textbf{66.19 ± 3.55}} & {\color[HTML]{3B3B3B} 65.79 ± 1.1} & {\color[HTML]{3B3B3B} {\ul 20.68 ± 4.08}} & {\color[HTML]{3B3B3B} \textbf{3.44 ± 1.09}} & {\color[HTML]{3B3B3B} 22.25 ± 0.36} & {\color[HTML]{3B3B3B} {\ul 28.35 ± 2.7}} & {\color[HTML]{3B3B3B} 38.98 ± 0.66} & {\color[HTML]{3B3B3B} \textbf{25.32 ± 0.42}} & 1463 \\
Test & {\color[HTML]{3B3B3B} sv} & {\color[HTML]{3B3B3B} \textbf{43.76 ± 4.23}} & {\color[HTML]{3B3B3B} \textbf{43.76 ± 4.23}} & {\color[HTML]{3B3B3B} 39.73 ± 1.35} & {\color[HTML]{3B3B3B} {\ul 27.61 ± 5.11}} & {\color[HTML]{3B3B3B} \textbf{6.12 ± 3.8}} & {\color[HTML]{3B3B3B} 28.89 ± 0.48} & {\color[HTML]{3B3B3B} {\ul 39.47 ± 4.33}} & {\color[HTML]{3B3B3B} \textbf{29.73 ± 0.59}} & {\color[HTML]{3B3B3B} 42.33 ± 1.81} & 112 \\
 & all & {\color[HTML]{3B3B3B} \textbf{64.22 ± 3.46}} & {\color[HTML]{3B3B3B} \textbf{64.22 ± 3.46}} & {\color[HTML]{3B3B3B} 63.7 ± 1.44} & {\color[HTML]{3B3B3B} {\ul 21.1 ± 4.14}} & {\color[HTML]{3B3B3B} \textbf{3.21 ± 1.4}} & {\color[HTML]{3B3B3B} 22.62 ± 0.39} & {\color[HTML]{3B3B3B} {\ul 29.15 ± 2.2}} & {\color[HTML]{3B3B3B} 38.32 ± 0.58} & {\color[HTML]{3B3B3B} \textbf{26.53 ± 0.32}} & 1575 \\ \hline
{\color[HTML]{3B3B3B} } & {\color[HTML]{3B3B3B} u} & {\color[HTML]{3B3B3B} 68.86 ± 1.52} & {\color[HTML]{3B3B3B} 68.86 ± 1.52} & {\color[HTML]{3B3B3B} \textbf{69.57 ± 0.77}} & {\color[HTML]{3B3B3B} {\ul 21.07 ± 4.72}} & {\color[HTML]{3B3B3B} \textbf{4.39 ± 1.28}} & {\color[HTML]{3B3B3B} 23.39 ± 0.62} & {\color[HTML]{3B3B3B} {\ul 26.85 ± 2.93}} & {\color[HTML]{3B3B3B} 38.11 ± 0.75} & {\color[HTML]{3B3B3B} \textbf{23.31 ± 0.72}} & 1381 \\
Dev & {\color[HTML]{3B3B3B} sv} & {\color[HTML]{3B3B3B} 41.33 ± 5.47} & {\color[HTML]{3B3B3B} 41.33 ± 5.47} & {\color[HTML]{3B3B3B} \textbf{47.39 ± 4.39}} & {\color[HTML]{3B3B3B} 22.03 ± 4.34} & {\color[HTML]{3B3B3B} \textbf{6.45 ± 1.95}} & {\color[HTML]{3B3B3B} {\ul 20.76 ± 1.49}} & {\color[HTML]{3B3B3B} 37.44 ± 2.59} & {\color[HTML]{3B3B3B} \textbf{26.74 ± 1.8}} & {\color[HTML]{3B3B3B} {\ul 34.53 ± 0.27}} & 135 \\
 & {\color[HTML]{3B3B3B} all} & {\color[HTML]{3B3B3B} 66.49 ± 2.19} & {\color[HTML]{3B3B3B} 66.49 ± 2.19} & {\color[HTML]{3B3B3B} \textbf{67.58 ± 0.53}} & {\color[HTML]{3B3B3B} {\ul \textit{20.99 ± 4.88}}} & {\color[HTML]{3B3B3B} \textbf{4.2 ± 1.47}} & {\color[HTML]{3B3B3B} 23.04 ± 0.47} & {\color[HTML]{3B3B3B} {\ul 27.79 ± 2.52}} & {\color[HTML]{3B3B3B} 37.09 ± 0.84} & {\color[HTML]{3B3B3B} \textbf{24.31 ± 0.64}} & 1516 \\ \hline
LegalBERT & {\color[HTML]{3B3B3B} u} & {\color[HTML]{3B3B3B} \textbf{69.3 ± 1.88}} & {\color[HTML]{3B3B3B} \textbf{69.3 ± 1.88}} & {\color[HTML]{3B3B3B} 67.31 ± 0.5} & {\color[HTML]{3B3B3B} 23.32 ± 1.01} & {\color[HTML]{3B3B3B} \textbf{2.95 ± 0.67}} & {\color[HTML]{3B3B3B} {\ul 22.04 ± 0.63}} & {\color[HTML]{3B3B3B} {\ul 25.1 ± 1.17}} & {\color[HTML]{3B3B3B} 37.23 ± 1.71} & {\color[HTML]{3B3B3B} \textbf{24.7 ± 0.17}} & 1463 \\
Test & {\color[HTML]{3B3B3B} sv} & {\color[HTML]{3B3B3B} \textbf{53.67 ± 4.59}} & {\color[HTML]{3B3B3B} \textbf{53.67 ± 4.59}} & {\color[HTML]{3B3B3B} 46.21 ± 4.41} & {\color[HTML]{3B3B3B} 29.92 ± 2.81} & {\color[HTML]{3B3B3B} \textbf{8.02 ± 0.58}} & {\color[HTML]{3B3B3B} {\ul 28.49 ± 0.82}} & {\color[HTML]{3B3B3B} 41.03 ± 2.75} & {\color[HTML]{3B3B3B} \textbf{28.28 ± 0.99}} & {\color[HTML]{3B3B3B} {\ul 40.72 ± 1.0}} & 112 \\
 & all & {\color[HTML]{3B3B3B} \textbf{68.03 ± 1.29}} & {\color[HTML]{3B3B3B} \textbf{68.03 ± 1.29}} & {\color[HTML]{3B3B3B} 65.16 ± 0.82} & {\color[HTML]{3B3B3B} 23.75 ± 1.08} & {\color[HTML]{3B3B3B} \textbf{2.99 ± 0.83}} & {\color[HTML]{3B3B3B} {\ul 22.32 ± 0.64}} & {\color[HTML]{3B3B3B} {\ul 26.23 ± 1.0}} & {\color[HTML]{3B3B3B} 36.6 ± 1.52} & {\color[HTML]{3B3B3B} \textbf{25.84 ± 0.21}} & 1575 \\ \hline
{\color[HTML]{3B3B3B} } & {\color[HTML]{3B3B3B} u} & {\color[HTML]{3B3B3B} \textbf{71.12 ± 0.67}} & {\color[HTML]{3B3B3B} \textbf{71.12 ± 0.67}} & {\color[HTML]{3B3B3B} 70.38 ± 0.67} & {\color[HTML]{3B3B3B} 24.69 ± 0.84} & {\color[HTML]{3B3B3B} \textbf{2.33 ± 1.65}} & {\color[HTML]{3B3B3B} {\ul 23.4 ± 0.75}} & {\color[HTML]{3B3B3B} {\ul 23.83 ± 0.29}} & {\color[HTML]{3B3B3B} 36.21 ± 1.57} & {\color[HTML]{3B3B3B} \textbf{22.9 ± 0.96}} & 1381 \\
Dev & {\color[HTML]{3B3B3B} sv} & {\color[HTML]{3B3B3B} 42.17 ± 2.92} & {\color[HTML]{3B3B3B} 42.17 ± 2.92} & {\color[HTML]{3B3B3B} \textbf{44.51 ± 3.23}} & {\color[HTML]{3B3B3B} {\ul 24.08 ± 0.96}} & {\color[HTML]{3B3B3B} \textbf{8.48 ± 2.47}} & {\color[HTML]{3B3B3B} 24.87 ± 1.92} & {\color[HTML]{3B3B3B} 40.91 ± 1.2} & {\color[HTML]{3B3B3B} \textbf{27.44 ± 0.81}} & {\color[HTML]{3B3B3B} {\ul 35.4 ± 1.63}} & 135 \\
 & {\color[HTML]{3B3B3B} all} & {\color[HTML]{3B3B3B} \textbf{68.29 ± 0.98}} & {\color[HTML]{3B3B3B} \textbf{68.29 ± 0.98}} & {\color[HTML]{3B3B3B} 68.28 ± 0.66} & {\color[HTML]{3B3B3B} 24.36 ± 0.97} & {\color[HTML]{3B3B3B} \textbf{2.35 ± 1.02}} & {\color[HTML]{3B3B3B} {\ul 23.26 ± 1.0}} & {\color[HTML]{3B3B3B} {\ul 25.35 ± 0.3}} & {\color[HTML]{3B3B3B} 35.43 ± 1.44} & {\color[HTML]{3B3B3B} \textbf{24.02 ± 0.75}} & 1516 \\ \hline
\end{tabular}

}       
        \caption{COC performance (hm-F1), confidence-calibration (ECE), and human-calibration (DistCE) performance with std($\pm$) in dev and test set. "/": COC finetuned models on \autoref{sec:coc}; "soft": models fine-tuned with soft-loss; "TS": model after Temperature Scaling. We choose the temperature $t$ by searching a range of possible values for $t$ on the dev set. We noted that the chosen $t$ across three random seeds were consistently overly aggressive, with values of 5.5, 5.8, and 5.5.
        Calibration results with standard deviation; Results shown over 3 random seeds.  }
        \label{tab:calibration_results_app}
    
    \end{table*}


\end{document}